\documentclass[runningheads]{llncs}
\usepackage{graphicx}
\usepackage{floatrow}
\usepackage[table]{xcolor}

\usepackage{tikz}
\usepackage{wrapfig}
\usepackage{comment}
\usepackage{multirow}
\usepackage{subcaption}
\usepackage{amsmath,amssymb} 
\usepackage{algorithm}
\usepackage{algpseudocode}
\usepackage{booktabs} 
\usepackage{diagbox}
\usepackage{color}
\usepackage{pifont}
\newcommand{\cmark}{\ding{51}}%
\newcommand{\xmark}{\ding{53}}%

\newcommand{\ie}{\textit{i}.\textit{e}., }
\newcommand{\eg}{\textit{e}.\textit{g}., }
\newfloatcommand{capbtabbox}{table}[][\FBwidth]

\usepackage[accsupp]{axessibility}  

\usepackage[width=122mm,left=12mm,paperwidth=146mm,height=193mm,top=12mm,paperheight=217mm]{geometry}
\usepackage[bookmarksnumbered=true]{hyperref} 

\hypersetup{
     colorlinks = true,
     linkcolor = black,
     anchorcolor = black,
     citecolor = blue,
     filecolor = black,
     urlcolor = blue
     }

\begin{document}

\pagestyle{headings}
\mainmatter
\author{Anonymous ECCV submission}
\def\ECCVSubNumber{3772}  

\title{Adversarial Detection without \\ Model Information} 

\institute{Paper ID \ECCVSubNumber}
\author{Abhishek Moitra, Youngeun Kim, and Priyadarshini Panda }
\institute{
Department of Electrical Engineering\\
Yale University\\
New Haven, CT, USA\\
{\tt\tiny \{abhishek.moitra, youngeun.kim, priya.panda\}@yale.edu}
}
\maketitle


\begin{abstract}
Prior state-of-the-art adversarial detection works are classifier model dependent, \ie they require classifier model outputs and parameters for training the detector or during adversarial detection. This makes their detection approach classifier model specific. Furthermore, classifier model outputs and parameters might not always be accessible. To this end, we propose a classifier model independent adversarial detection method using a simple energy function to distinguish between adversarial and natural inputs. We train a standalone detector independent of the classifier model, with a layer-wise energy separation (LES) training to increase the separation between natural and adversarial energies. With this, we perform energy distribution-based adversarial detection. Our method achieves comparable performance with state-of-the-art detection works (ROC-AUC $>$ 0.9) across a wide range of gradient, score and gaussian noise attacks on CIFAR10, CIFAR100 and TinyImagenet datasets. Furthermore, compared to prior works, our detection approach is light-weight, requires less amount of training data (40\% of the actual dataset) and is transferable across different datasets. For reproducibility, we provide layer-wise energy separation training code at \href{https://github.com/Intelligent-Computing-Lab-Yale/Energy-Separation-Training}{https://github.com/Intelligent-Computing-Lab-Yale/Energy-Separation-Training}
\end{abstract}

\section{Introduction}
\label{sec:intro}
Deep Neural Networks (DNNs) are vulnerable to adversarial attacks \cite{akhtar2018threat,papernot2017practical} where
small crafted noise is added to natural images to fool a classifier model causing
high confidence mis-classifications. Prior adversarial defense works have focused
on improving the prediction of classifier models on adversarial inputs. These works have used innovative techniques such as adversarial training \cite{madry2017towards}, input transformation \cite{guo2017countering}, randomization \cite{xie2017mitigating} among others. Recently, Adversarial detection has emerged as a strong defence strategy against adversarial attacks. Here, a “detector” network is trained to identify adversarial
and natural inputs in a system \cite{metzen2017detecting,yin2019gat,xu2017feature}. 
\begin{figure}

  \includegraphics[width=0.8\textwidth]{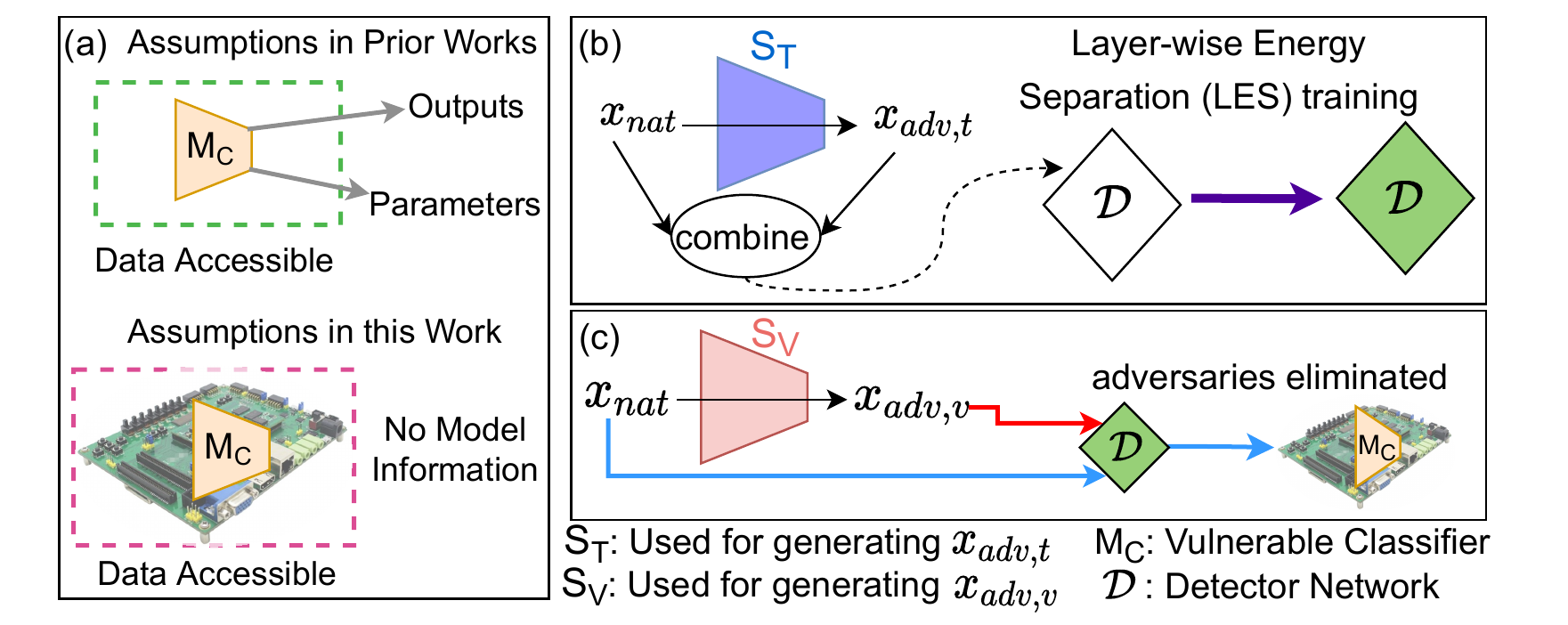}
  \caption{(a) In our proposed approach, we assume that the classifier model's ($M_C$) information (output and parameter) is in-accessible for creating the adversarial detector or during adversarial detection. However, we do require access to the dataset during detector training. (b) The detector $\mathcal{D}$ is trained on natural ($x_{nat}$) and adversarial ($x_{adv,t}$) data using a layer-wise energy separation (LES) training. The $x_{adv,t}$ data is generated using $x_{nat}$ and model $S_T$. (c) The LES-trained detector can identify adversaries generated using a different model $S_V$. Note, $S_T$ and $S_V$ can have different network architectures from the classifier model $M_C$.}%
  \label{proposed_method}
\vspace{-3mm}
\end{figure}
Prior adversarial detection works heavily rely on the classifier model outputs and parameters for detection. However, the reliance on classifier model has two major drawbacks: 1) it reduces the generalizability of adversarial detectors. This means that detector networks are specific to each classifier model and need to be retrained if the classifier model changes; 2) classifier model parameters and outputs might be in-accessible \cite{umuroglu2017finn,george2018efficient,yin2020xnor}. To this effect, we propose an adversarial detection approach with the following assumption: The outputs and parameters of an adversarially vulnerable classifier model $M_C$ are in-accessible for adversarial detection. In other words, \textbf{no classifier model information is required for creating the detector or during adversarial detection}. Fig. \ref{proposed_method}a compares the assumptions in this work with prior adversarial detection works. \textbf{Note, although our work does not require classifier model information we do require data accessibility for the detection.}

In this work, we train a small, standalone detector network $\mathcal{D}$ without any information about $M_C$. Firstly, we propose a simple energy function to map a high dimensional input feature space to a one-dimensional output feature space (called the energy). As adversaries are created by adding noise to natural data, we find that the energies corresponding to natural and adversarial data are different. However, the difference is not sufficient to perform reliable adversarial detection. To this end, we employ a layer-wise energy separation (LES) training using an energy distance-based loss function to maximize the separation between natural and adversarial energies.

\begin{table*}[t]
    \addtolength{\tabcolsep}{2.5pt}
    \centering
\caption{Table showing the key differences between prior adversraial detection works and our proposed method (\cmark: addressed / \xmark: not addressed, \textit{G}: Gradient-based, \textit{S}: Score-based and \textit{GN}: Gaussian noise attacks). Note, A $\rightarrow$ B signifies that a detector trained on A-type attacks can successfully detect B-type attacks.}
\label{tab:prelim_comparison}
\resizebox{1\textwidth}{!}{%
\begin{tabular}{lcccc}
\hline
\multirow{2}{*}{Work} &  \multicolumn{2}{c}{Transferability Across} & \multirow{2}{*}{\begin{tabular}[c]{@{}c@{}}Dataset\\Access\end{tabular}} & \multirow{2}{*}{\begin{tabular}[c]{@{}c@{}}Classifier Model \\ Access Required?\end{tabular}} \\ \cline{2-3}
 & Attacks & Datasets &  \\ \hline
\begin{tabular}[c]{@{}l@{}}Metzen et al. \cite{metzen2017detecting}, Yin et al. \cite{yin2019gat}, \\Sterneck et al. \cite{sterneck2021noise}\end{tabular} & G $\rightarrow$ G & \begin{tabular}[c]{@{}c@{}}\xmark\end{tabular} & Full Access & \begin{tabular}[c]{@{}c@{}}intermediate\\ activations \end{tabular} \\ \hline
\begin{tabular}[c]{@{}l@{}}Moitra et al. \cite{moitra2021detectx}, Xu et al. \cite{xu2017feature},\\ Grosse et al. \cite{grosse2017statistical}\end{tabular} & G $\rightarrow$ G & \begin{tabular}[c]{@{}c@{}}\xmark \end{tabular} & Full Access & \begin{tabular}[c]{@{}c@{}}model\\ training \end{tabular} \\ \hline
Huang et al. \cite{huang2019model} & G $\rightarrow$ G & \xmark & Full Access & model outputs \\ \hline
\textbf{Ours} & \textbf{G} $\rightarrow$ \textbf{G},  \textbf{G} $\rightarrow$ \textbf{S}, \textbf{G} $\rightarrow$ \textbf{GN} & \textbf{\cmark} & \textbf{Partial Access} & \textbf{No access} \\ \hline
\end{tabular}%
}
\vspace{-3mm}
\end{table*}

In Fig. \ref{proposed_method}b it can be seen that the detector $\mathcal{D}$ is trained on natural ($x_{nat}$) and adversarial ($x_{adv,t}$) data. Here, $x_{adv,t}$ is created using model $S_{T}$ that is different from $M_C$. Interestingly, we find that the LES-trained detector can identify attacks generated from the model $S_V$ that is different from both $S_{T}$ and $M_C$. We perform extensive analysis on a wide range of gradient \cite{madry2017towards,carlini2017towards,dong2018boosting}, score \cite{andriushchenko2020square,croce2020reliable} and decision based \cite{croce2020minimally} attacks on datasets such as CIFAR10, CIFAR100 and TinyImagenet. Through our experiments, we discover certain interesting features of our detection approach that have not been shown in prior works \cite{gong2017adversarial,metzen2017detecting,yin2019gat,moitra2021detectx}. \ding{202} Firstly, our approach is agnostic to the model used for generating attacks. For example, a detector trained using $S_{T}$=VGG16 can identify attacks generated using $S_V$=ResNet18. \ding{203} Compared to prior works that have shown transferability on gradient to gradient-based attacks only, our detection approach can transfer across different types of adversarial attacks. For example, a detector trained on gradient-based attacks can detect certain score-based attacks such as Square \cite{andriushchenko2020square}, Auto-PGD \cite{croce2020reliable}. \ding{204} Additionally, through our ablation studies, we find that the error amplification effect discussed in \cite{lin2019defensive} aids in improving the energy separation of our detector. Consequently, we find that a higher Lipschitz constant \cite{cisse2017parseval} enables better adversarial detection. 

As mentioned earlier, our detection approach requires access to the data. To this end, we evaluate the performance of our detection approach with lower access to the data. \ding{202} We find that the LES training can achieve significantly high performance even with limited data accessibility (\eg with 40\% access to the dataset). \ding{203} We also find that the detector is transferable across different datasets. For instance, a detector constructed using CIFAR100 data can detect adversarial attacks from the CIFAR10 and TinyImagenet datasets. However, a small amount (about 200 data samples) of the CIFAR10 and TinyImagenet datasets is still required for the transfer. Table \ref{tab:prelim_comparison} compares the salient features of our detection work with prior state-of-the-art adversarial detectors. 

In summary, the contributions of our work are:
\begin{itemize}
    \item We propose a simple and light-weight energy metric to distinguish between clean and adversarial inputs. To perform adversarial detection, we maximize the energy distance between them using an energy distance-based objective function and a layer-wise energy separation (LES) training approach. Our adversarial detection approach does not require any access to the classifier model parameters or outputs. 
    \item We perform extensive experiments on benchmark datasets like CIFAR10, CIFAR100, and TinyImagenet with state-of-the-art gradient-based \cite{madry2017towards,kurakin2016adversarial}, score-based \cite{croce2020reliable} and decision-based \cite{croce2020minimally} adversarial attacks. We find our approach yields state-of-the-art detection across different adversarial attacks and datasets. 
    \item Through our experiments, we find that our detection approach is agnostic to the model used for attack generation. Additionally, we show that the detector trained on gradient-based attacks can detect certain score-based attacks and gaussian noise attacks. 
    \item Through extensive experiments, we show that the LES training can achieve high performance with limited access to the dataset. Further, our adversarial detector can transfer across different datasets. Using transferability, a detector trained on one dataset can be reused to detect adversaries created on another dataset.
\end{itemize}


\section{Background}
\subsection{Adversarial Attacks}
In this work, we consider three different kinds of adversarial attacks. Here, we discuss those attacks in detail. 

\noindent\textbf{Gradient-based attacks:} These attacks require gradients of the input to craft adversaries. The Fast Gradient Sign Method (FGSM) is a simple one-step adversarial attack proposed in \cite{kurakin2016adversarial}. Several works have shown that FGSM attack can be made stronger with momentum (MIFGSM) \cite{dong2018boosting}, random initialization (FFGSM) \cite{wong2020fast}, and input diversification (DIFGSM) \cite{xie2019improving}. In contrast, the Basic Iterative Method (BIM) is an iterative attack proposed in \cite{kurakin2016adversarial}. The BIM attack with random restarts is called the Projected Gradient Descent (PGD) attack \cite{madry2017towards}. A targeted version of the PGD attack (TPGD) \cite{zhang2019theoretically} can fool the model into mis-classifying a data as a desired class. Other multi-step attacks like Carlini-Wagner (C\&W) \cite{carlini2017towards} and PGD-L2 \cite{madry2017towards} are crafted by computing the L2 Norm distance between the adversary and the natural images.


\noindent\textbf{Score-based attacks:} Score-based attacks do not require input gradients to craft adversaries. Square attack (SQR) \cite{andriushchenko2020square} uses multiple queries to perturb randomly selected square regions in the input. Other score-based attacks like the Autoattack (AUTO) and Auto-PGD (APGD) craft adversaries by automatically choosing the optimal attack parameters \cite{croce2020reliable}.

\noindent\textbf{Decision-based attacks:} These attacks craft adversaries based on the decision outputs of a model. The Fast Adaptive Boundary (FAB) \cite{croce2020minimally} attack finds the minimum perturbation required to perform a mis-classification. The Gaussian Noise (GN) attack is created by adding gaussian noise to the input. 

\subsection{Performance Metrics for Adversarial Detection}
\label{auc_error_accuracy}
To evaluate the adversarial detection performance, we use three metrics: {ROC-AUC, Accuracy} and {Error}. 

\noindent\textbf{Area Under the ROC Curve (AUC Score):} The area under the ROC curve, compares the True Positive Rate (TPR) and the False Positive Rate (FPR) of a classifier. A high ROC-AUC score signifies a good classifier \cite{yin2019gat}.

\noindent\textbf{Accuracy and Error:} In this work, \textit{Accuracy} is defined as the fraction of natural inputs that are correctly classified by the classifier model and not rejected by the adversarial detector. While, {\it{Error}} is defined as the fraction of adversarial inputs that are classified incorrectly by the classifier model and not rejected by the detector. Thus, high accuracy and low error is desirable.
\section{Related Works}

\subsection{Adversarial Classification}
\label{classification_works}
Here, the objective is to improve the adversarial classification accuracy of a vulnerable model. Guo et al. \cite{guo2017countering} proposed input feature transformation using JPEG compression followed by training on compressed feature space to improve the classification performance of the classifier model. Madry et al. \cite{madry2017towards} proposed adversarial training in which a classifier model is trained on adversarial and clean data to improve the adversarial and clean classification performance. Following this, several works have used noise injection into parameters \cite{he2019parametric} and ensemble adversarial training to harden the classifier model against a wide range of attacks. Lin et al. \cite{lin2019defensive} showed that adversarial classification can be improved by reducing the error amplification in a network. Hence, they used adversarial training with regularization to constrain the Lipschitz constant of the network to less than unity. In our work, we do not improve the adversarial classification accuracy of the classifier model. Rather, we focus on adversarial detection to distinguish adversarial data from natural ones. 
\subsection{Adversarial Detection}
\subsubsection{Works requiring {classifier model} training}
\label{rel_works_training}
The following works require training of the classifier model to perform adversarial detection. Xu et al. \cite{xu2017feature} propose a method that uses outputs of multiple classifier models to estimate the difference between natural and adversarial data. Here, the classifier models are trained on natural inputs with different feature squeezing techniques at the inputs. Moitra et al. \cite{moitra2021detectx} uses the features from the first layer of the underlying model to perform adversarial detection. In particular, they perform adversarial detection using hardware signatures in DNN accelerators. Further, several recent works like \cite{gong2017adversarial} have shown that adversarial and natural data have different data distributions. While Grosse et al. \cite{grosse2017statistical} train the classifier model with an additional class label indicating adversarial data, Gong et al. \cite{gong2017adversarial} train a separate binary classifier on the natural and adversarial data generated from the classifier model to perform adversarial detection. Lee et al. \cite{lee2018simple} use a metric called the \textit{Mahalanobis distance} classifier to train the classifier model. The Mahalanobis distance is used to distinguish natural from adversarial data.

\subsubsection{Works requiring {classifier model} outputs}
\label{rel_works_outputs}
These works show that adversarial and natural inputs can be distinguished based on the intermediate features of the classifier model. Metzen et al. \cite{metzen2017detecting} and Sterneck et al. \cite{sterneck2021noise} use the intermediate features to train a simple binary classifier for adversarial detection. While Metzen et al. \cite{metzen2017detecting} use a heuristic-based method to determine the point of attachment of the detector with the classifier model, Sterneck et al. \cite{sterneck2021noise} use a structured metric called adversarial noise sensitivity to do the same. Similarly, Yin et al. \cite{yin2019gat} use asymmetric adversarial training to train detectors on the intermediate features of the classifier model for adversarial detection. Another work by Ahuja et al. \cite{ahuja2019probabilistic} use the data distributions from the intermediate layers in the classifier model and Gaussian mixture models to perform adversarial detection. Further, Huang et al. \cite{huang2019model} use the confidence scores from the classifier model to estimate the \textit{relative score difference} corresponding to the clean and the adversarial input to perform adversarial detection. Further, they also recommend classifier model training on noisy data to improve the adversarial detection performance.   

Evidently, prior works discussed in the Section \ref{classification_works} and Section \ref{rel_works_outputs} require training or the outputs of the classifier model for adversarial detection. In contrast, our work is focused towards performing adversarial detection without accessing the classifier model.
\section{Layer-wise Energy Separation (LES) Training}
\label{methodology}
\subsection{Training Methodology}
In this work, we define a simple energy function for layer $l$, $\mathcal{E}^l$ shown in Eq. \ref{snm}. The energy is defined as the average magnitude of the feature outputs in a particular layer $l$ with outputs $Z^{c,h,w}_l$. Here $c, h,$ and $w$ are the number of output channels, height and width, respectively of the feature outputs in the layer $l$. In LES training, we train the detector network to maximize the energy separation between $\mathcal{E}_{nat}$ and $\mathcal{E}_{adv}$. Here, $\mathcal{E}_{nat}$ and $\mathcal{E}_{adv}$ are the energies corresponding to natural and adversarial inputs, respectively. Note, the energy separation is the difference between the mean values of $\mathcal{E}_{nat}$ and $\mathcal{E}_{adv}$ distributions.
\begin{equation}
    \mathcal{E}^l = \frac{1}{c h w}\sum_{i=1}^c\sum_{j=1}^h\sum_{k=1}^w |Z^{i,j,k}_l|.
    \label{snm}
\end{equation}
\textbf{Training objective:} 
We train the detector using the following objective function:
\begin{equation}
    \max_{\theta} ||\mathcal{E}_{nat}^l - \mathcal{E}_{adv}^l||,
    \label{objective}
\end{equation}
corresponding to any layer $l$. Here, $\theta$ denotes the parameters of the detector network. In order to achieve this, we design an energy separation-based loss function that minimizes the energy distances between $\mathcal{E}_{nat}^l$ ($\mathcal{E}_{adv}^l$) and $\lambda_{n}$ ($\lambda_{a}$). Here, $\lambda_{n}$ and $\lambda_{a}$ are hyper-parameters denoting the desired natural and adversarial energies, respectively. An indicator variable $y$ has the value of 1 or 0 for natural and adversarial inputs, respectively. 
\begin{equation}
    \mathcal{L} = y ~\mathcal{L}_{MSE}(\mathcal{E}_{nat}^l, \lambda_{n}) + (1-y) ~\mathcal{L}_{MSE}(\mathcal{E}_{adv}^l, \lambda_{a}).
    \label{loss}
\end{equation}
Note, our training objective is loosely based on the current signature separation in recent work \cite{moitra2021detectx}. We would like to highlight that \cite{moitra2021detectx} focuses on modifying the classifier model's layers to perform adversary detection. In contrast, we train the detector independently of the classifier model that enables high detection scores along with transferability across multiple datasets and attacks.

\textbf{Dataset generation:} The training dataset for the detector contains equal number of natural ($x_{nat}$) and adversarial data ($x_{adv,t}$). Here, $x_{adv,t}$ is generated using the model $S_{T}$ having a different network architecture than $M_C$ and trained on $x_{nat}$ using the stochastic gradient descent (SGD) algorithm.
\begin{algorithm}[t]\small

\algrenewcommand\algorithmicrequire{\textbf{Input:}}
\algrenewcommand\algorithmicensure{\textbf{Output:}}

\begin{algorithmic}[1]
\caption{Layer-wise Energy Separation (LES) training algorithm}
\Require $n$ layered detector ($\mathcal{D}$), $x_{nat}$, $x_{adv,t}$, $s_{nat}$
\Ensure $n$ layered trained detector $\mathcal{D}_T$, $\mathcal{E}_{Th}$
    
    \ForAll {i = 1 to n}
        \ForAll {j = 1 to $N_{epoch}$}
            \State {Freeze layers $[0,i$-$1]$}
            \State {Fetch mini-batch $X_n$ and $X_a$ from $x_{nat}$ and $x_{adv,t}$, respectively}
            \State {Compute $\mathcal{E}_{nat}^i$ and $\mathcal{E}_{adv}^i$ on the mini-batch}
            \State {Compute loss function using Eq. \ref{loss}}
            \State {Optimize layer $i$ using the loss function}
        \EndFor 
    \EndFor \\
Generate distribution $\mathcal{E}_{s_{nat}}$ with $s_{nat}$ and $\mathcal{D}_T$\\
$\mathcal{E}_{Th}$ $\gets$ $K^{th}$ percentile of $\mathcal{E}_{s_{nat}}$
\label{alg:algorithm}
\end{algorithmic}

\end{algorithm}

\begin{wrapfigure}{h}{0.4\textwidth}
    \vspace{-9mm}
    \begin{subfigure}[b]{\textwidth}
         \centering
         \includegraphics[width=\textwidth]{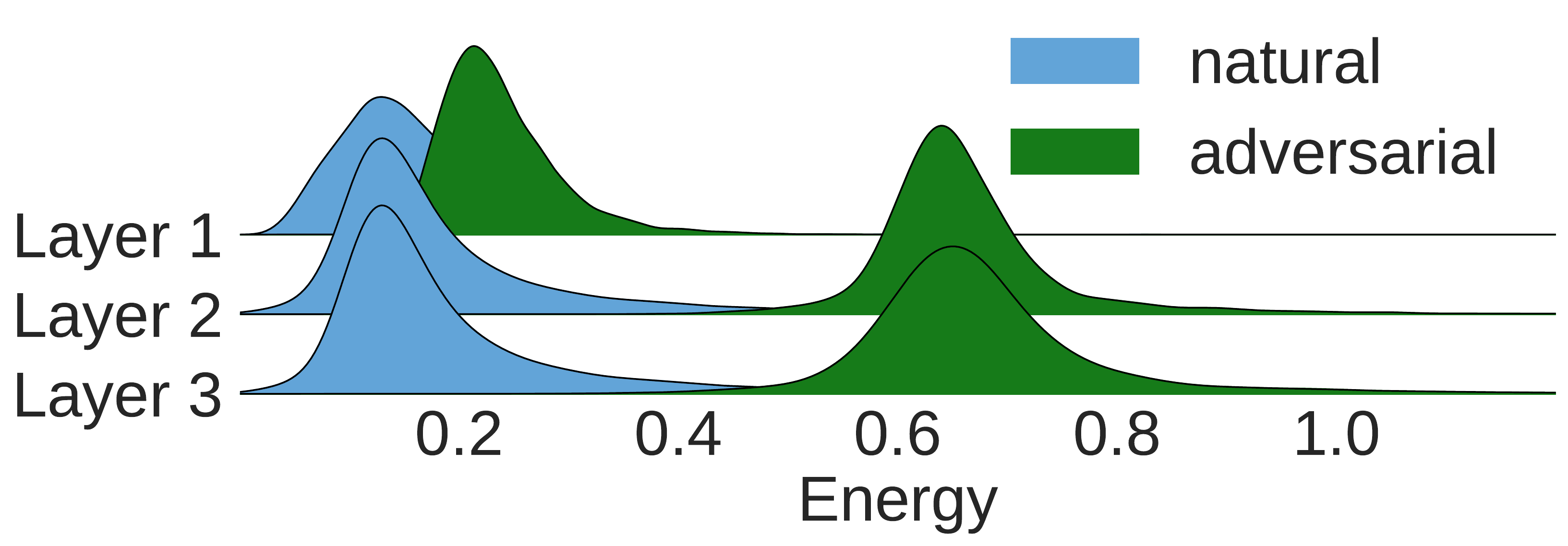}
         \caption{}
         \label{lwise_pgd}
     \end{subfigure}
     \\
     \begin{subfigure}[b]{\textwidth}
         \centering
         \includegraphics[width=\textwidth]{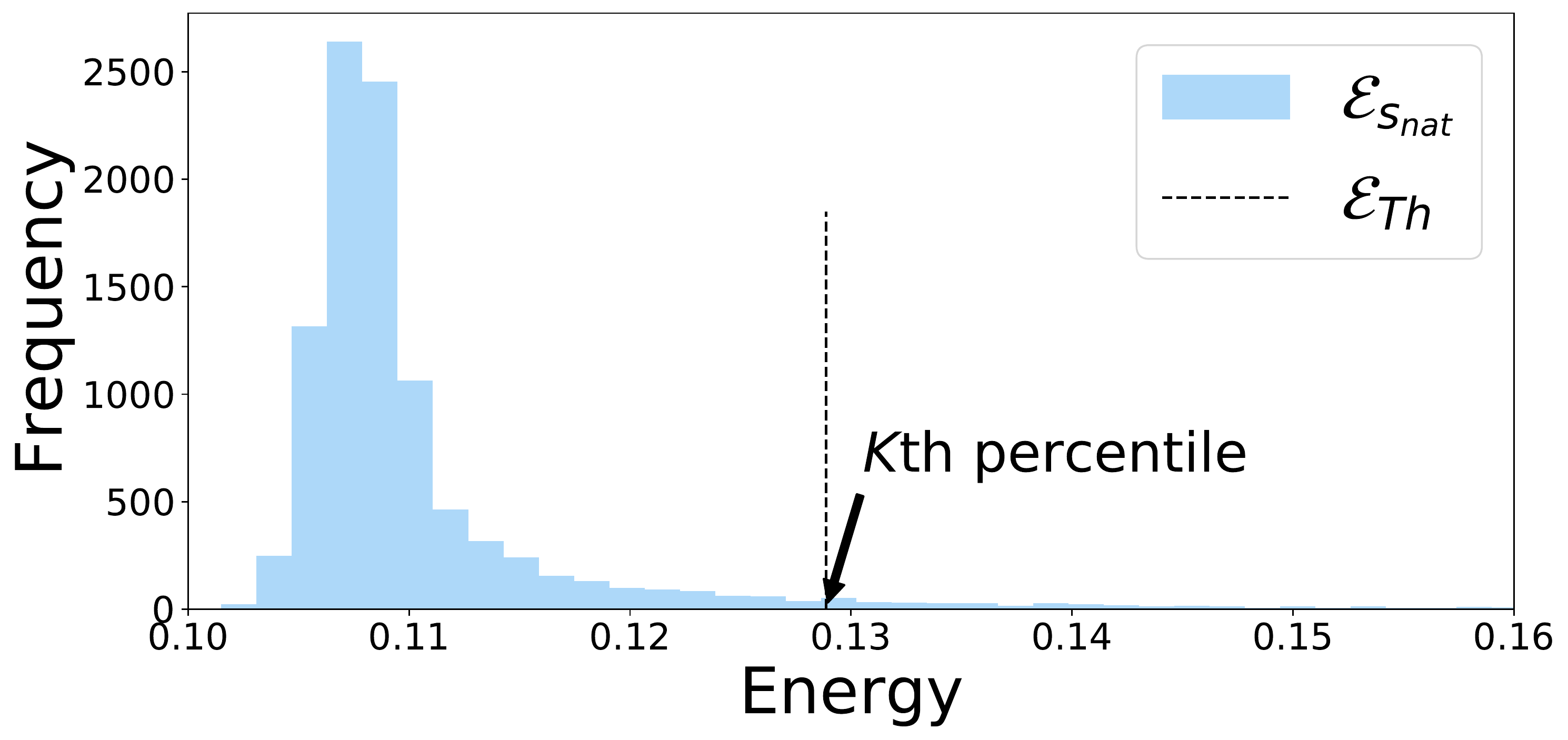}
         \caption{}
         \label{eth_select}
     \end{subfigure}
    
    \caption{a) The separation between the clean and adversarial energy distributions increase with with each layer as the LES training proceeds. b) The value of $\mathcal{E}_{Th}$ is set as the Kth percentile value of the $\mathcal{E}_{s_{nat}}$ distribution.}
    \label{lwise_training}
    \vspace{-5mm}
\end{wrapfigure}

\textbf{LES training methodology:} Algorithm \ref{alg:algorithm} shows the LES training approach. LES training begins with a randomly initialized $n$ layered detector, with $n$ being a hyper-parameter. Other inputs include the natural samples $x_{nat}$, adversarial samples $x_{adv,t}$ and $s_{nat}$. $s_{nat}$ is created by randomly selecting 200 data samples from $x_{nat}$. 

The training is carried out in multiple stages. In each stage $i$, layers [0,$i$-$1$] are frozen and layer $i$ is optimized for energy distance maximization. In each stage, mini-batches $X_n$ and $X_a$ are fetched from $x_{nat}$ and $x_{adv,t}$, respectively. Following this, the natural and adversarial energies for layer $i$ are computed. The energies are used to compute the loss shown in Eq. \ref{loss} which is then used to optimize the $i$th layer parameters of the detector. As layers [0, $i$-$1$] are frozen, the energy separation obtained till layer $i$-$1$ is preserved. Thus, training the $i${th} layer further increases the energy separation.

After the LES training, we obtain the trained detector $\mathcal{D}_T$. At this step, we use $s_{nat}$ and $\mathcal{D}_T$ to obtain the a sample natural energy distribution $\mathcal{E}_{s_{nat}}$. The $\mathcal{E}_{s_{nat}}$ is obtained at the final layer of $\mathcal{D}_T$. Next, the $K$th percentile of the $\mathcal{E}_{s_{nat}}$ distribution is chosen as the threshold energy $\mathcal{E}_{Th}$. All energy values greater than $\mathcal{E}_{Th}$ are classified as adversarial inputs and vice-versa.

\textbf{Demonstrating the Efficacy of LES training and Choosing $\mathcal{E}_{Th}$:} Fig. \ref{lwise_pgd} shows the energy separation between $\mathcal{E}_{nat}$ and $\mathcal{E}_{adv}$ obtained after each layer of a three-layered detector trained on the CIFAR100 dataset in the layer-wise manner. The adversarial inputs correspond to PGD(4,2,10)\footnote{PGD(4,2,10)= PGD with parameters $\epsilon$: 4/255, $\alpha$: 2/255, $steps$: 10.} attack on the CIFAR100 dataset. It can be seen that the energy separation increases with each layer in the detector. The increasing separation significantly improves the detection of weak attacks\footnote{Weak attacks have smaller $\epsilon$ values compared to stronger attacks} (like PGD(4,2,10)), that may go undetected at early layers due to low energy separation.

In the final layer, such attacks can be detected successfully. Fig. \ref{eth_select} shows the $\mathcal{E}_{s_{nat}}$ distribution obtained at the final layer of the trained detector $\mathcal{D}_T$. The value of $\mathcal{E}_{Th}$ is chosen as the $K$th percentile of the distribution.

\subsection{Lipschitz Constant and Energy Separation}
\begin{wrapfigure}{h}{0.38\textwidth}
    \vspace{-9mm}
    \centering
    \includegraphics[width=0.90\linewidth]{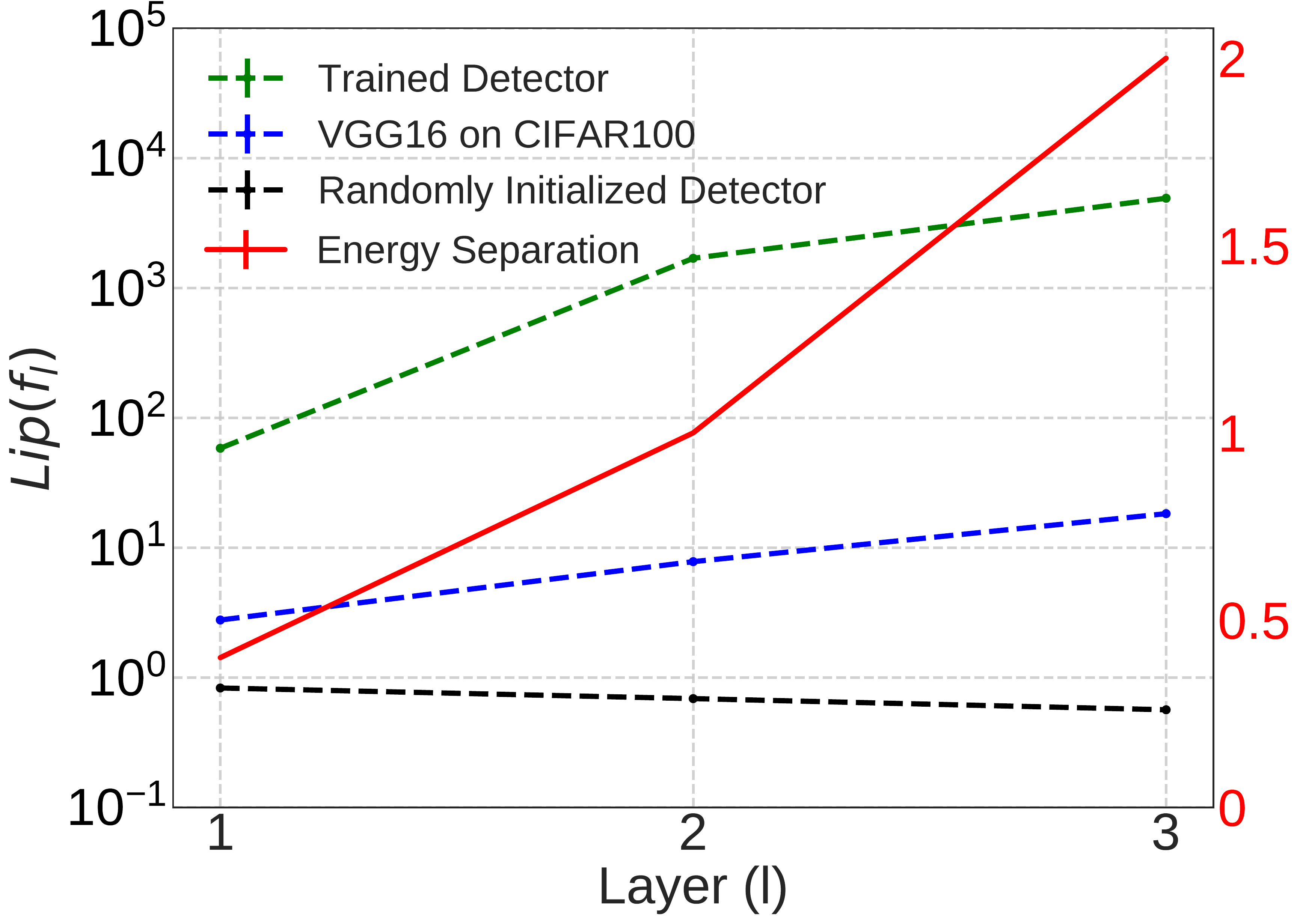}  
       \vspace{25mm}
    \caption{Comparison of Lipschitz constant of feed-forward function ($Lip(f_l)$) for three different networks. Note, only first three layers of the VGG16 network have been shown. It is observed that $Lip(f_l)$ increases exponentially as the energy separation (in red) between $\mathcal{E}_{nat}$ and $\mathcal{E}_{adv}$ increases. Compared to the detector, rise in $Lip(f_l)$ is negligible for the other two networks.}
    \label{lip_constant}
    \vspace{-27mm}
\end{wrapfigure}
The Lipschitz constant has been associated with the error amplification effect in a neural network \cite{lin2019defensive}. The Lipschitz constant of a feed-forward function $f_l$ for any layer $l$, can be expressed as follows:
\begin{equation}
    Lip(f_l) \leq \prod_{i=1}^l {Lip(\phi_i)}.
\end{equation}
where $\phi$ can be a linear, convolutional, max pooling or an activation layer. 

Fig. \ref{lip_constant} shows the $Lip(f_l)$ values for three different networks: i) A randomly initialized 3-layered detector network, ii) the 3-layered detector after LES training on the CIFAR100 dataset, and iii) a VGG16 network trained on the CIFAR100 dataset using SGD. Interestingly, we observe that the value of $Lip(f_l)$ rises as the energy separation between $\mathcal{E}_{nat}^l$ and $\mathcal{E}_{adv}^l$ increases. This suggests that the layer-wise training increases the error amplification factor in the detector network. 

In prior works that minimize the adversarial perturbations in a network \cite{cisse2017parseval,lin2019defensive}, it has been shown that a lower Lipschitz constant helps in reducing the adversarial perturbations and thus increase adversarial robustness. \textit{However, in this work, we show that a higher Lipschitz constant amplifies the energy separation and thus, higher adversarial detection.} 

\subsection{Effect of Detector Width}\vspace{-2mm}
\begin{wrapfigure}{h}{0.38\textwidth}
    \vspace{-10mm}
    \centering
    \includegraphics[width=1.05\linewidth]{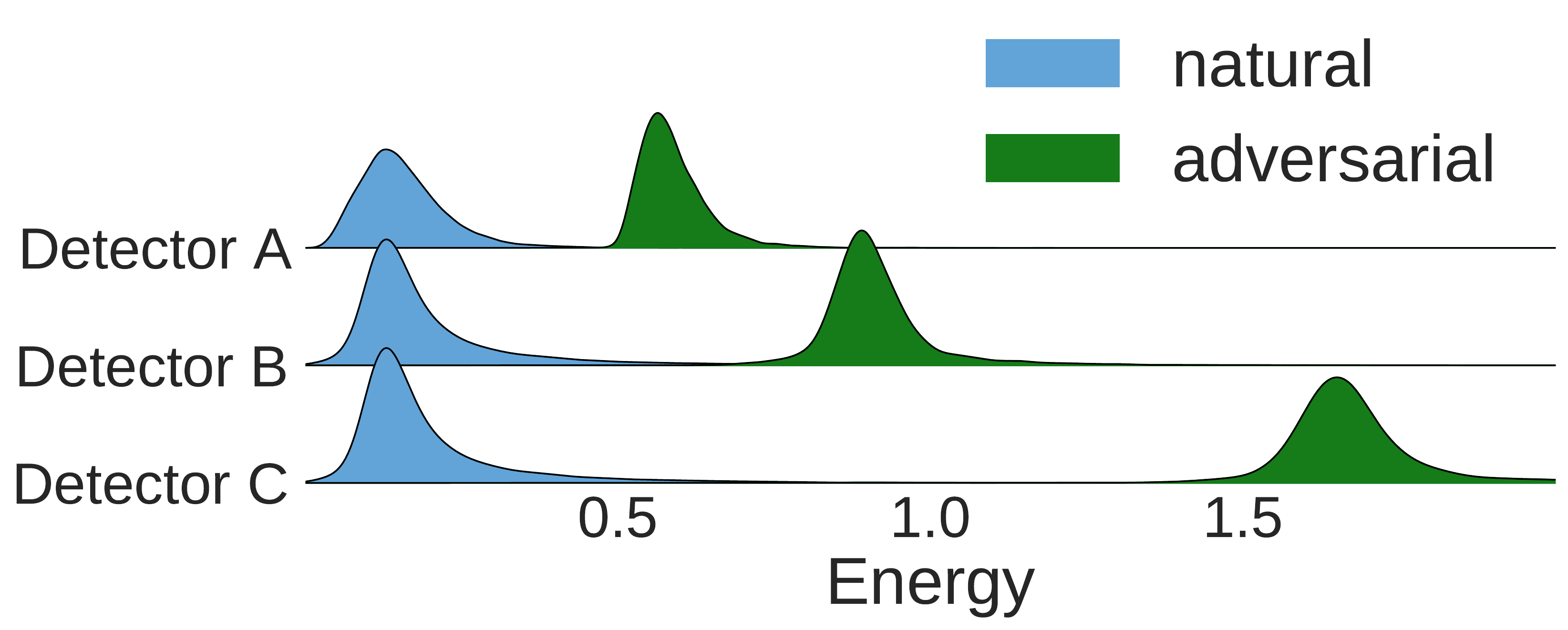}  
    \vspace{2mm}
    \caption{The clean and adversarial energy distributions corresponding to wide and narrow detector network architectures. Wider detector networks (detector A) achieve smaller energy separation compared to narrow detectors (detector C).}
    \label{arch_size}
        \vspace{-10mm}
\end{wrapfigure}
Fig. \ref{arch_size} shows the separation between natural and adversarial energies (corresponding to PGD(8,4,10) attacks) computed at the last layer of three detector models: \textit{Detector A}: Conv(3,64)-ReLU-Conv(64,128)- ReLU-Conv(128,256), \textit{Detector B}: Conv(3,32)-ReLU-\\ Conv(32,64)-ReLU-Conv(64,128) and \textit{Detector C}: Conv(3,8)-ReLU-Conv(8,16)-ReLU-\\Conv(16,32). Clearly, detectors having narrower layers achieve higher energy separation in the final layer compared to wider detector networks. \textit{Consequently, light-weight detectors improve the adversarial detection performance.} Further, we observe that detectors without batch normalization achieve higher energy separation compared to detectors with batch normalization layers. In all our experiments throughout the paper, we use detectors without batch-normalization layers. \vspace{-3mm}
\begin{figure}
    \begin{subfigure}[b]{0.3\textwidth}
         \centering
         \includegraphics[width=\textwidth]{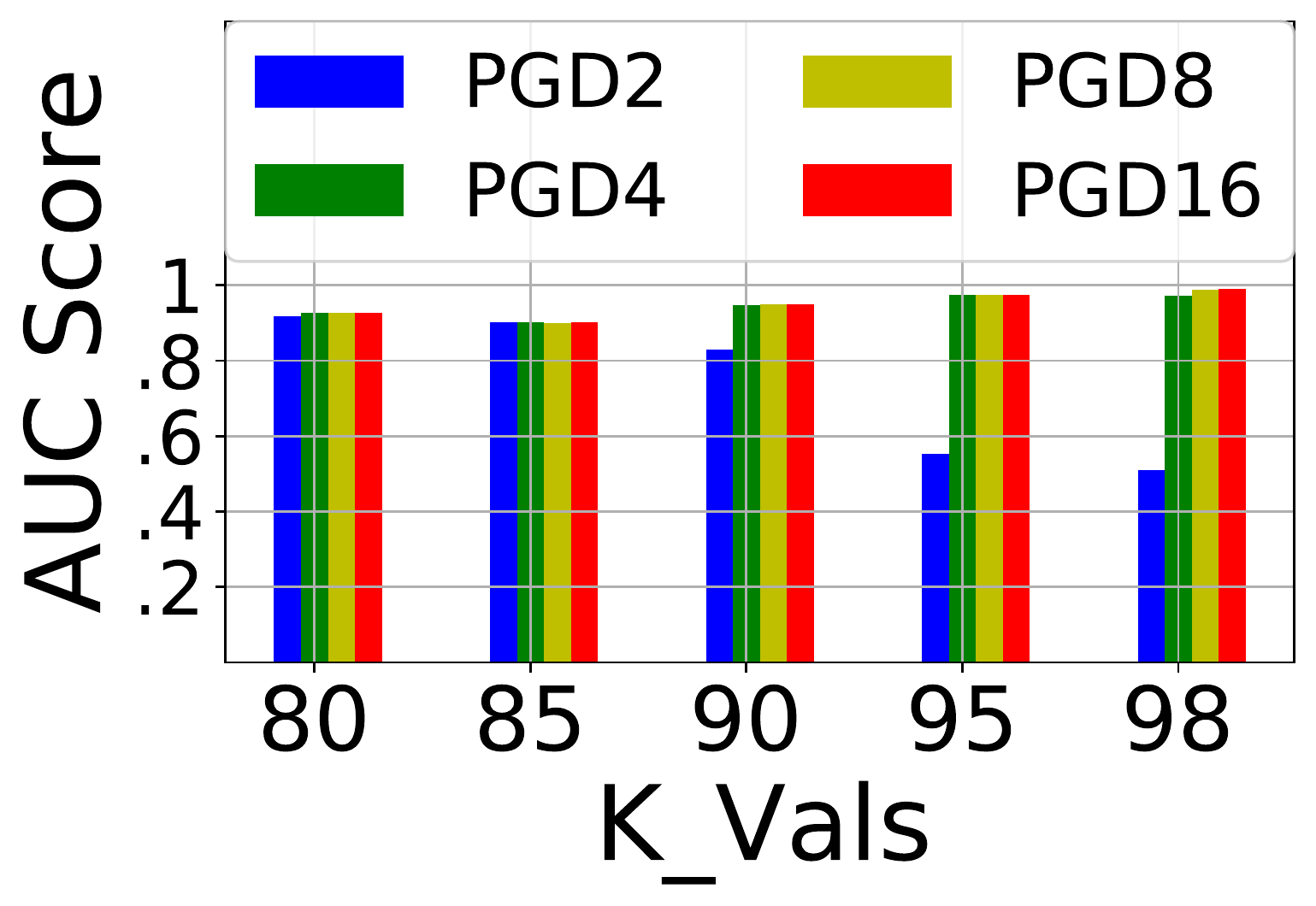}
         \caption{}
         \label{K_select}
     \end{subfigure}
     \begin{subfigure}[b]{0.3\textwidth}
         \centering
         \includegraphics[width=\textwidth]{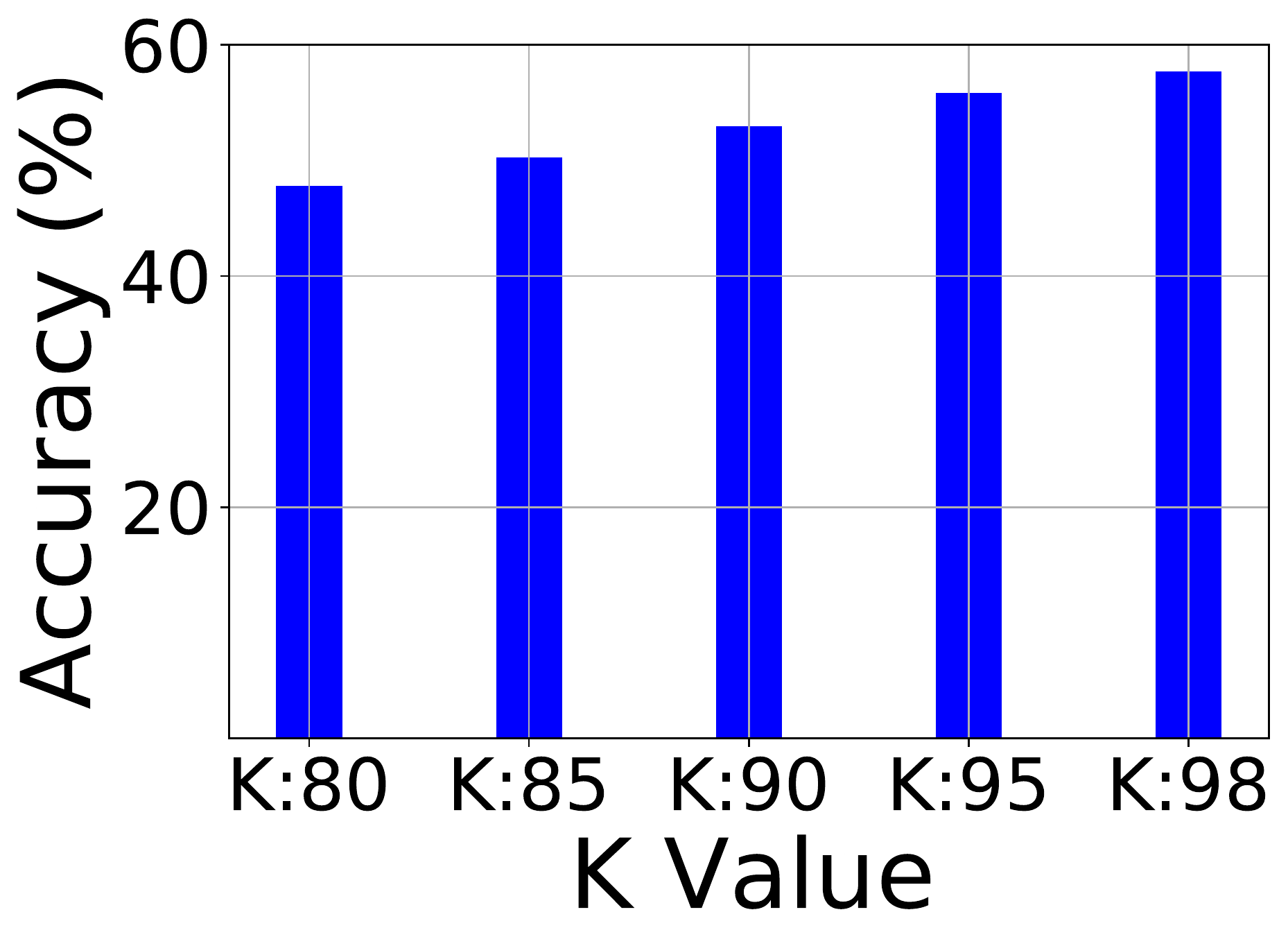}
         \caption{}
         \label{K_select_error}
     \end{subfigure}
    \begin{subfigure}[b]{0.3\textwidth}
         \centering
         \includegraphics[width=\textwidth]{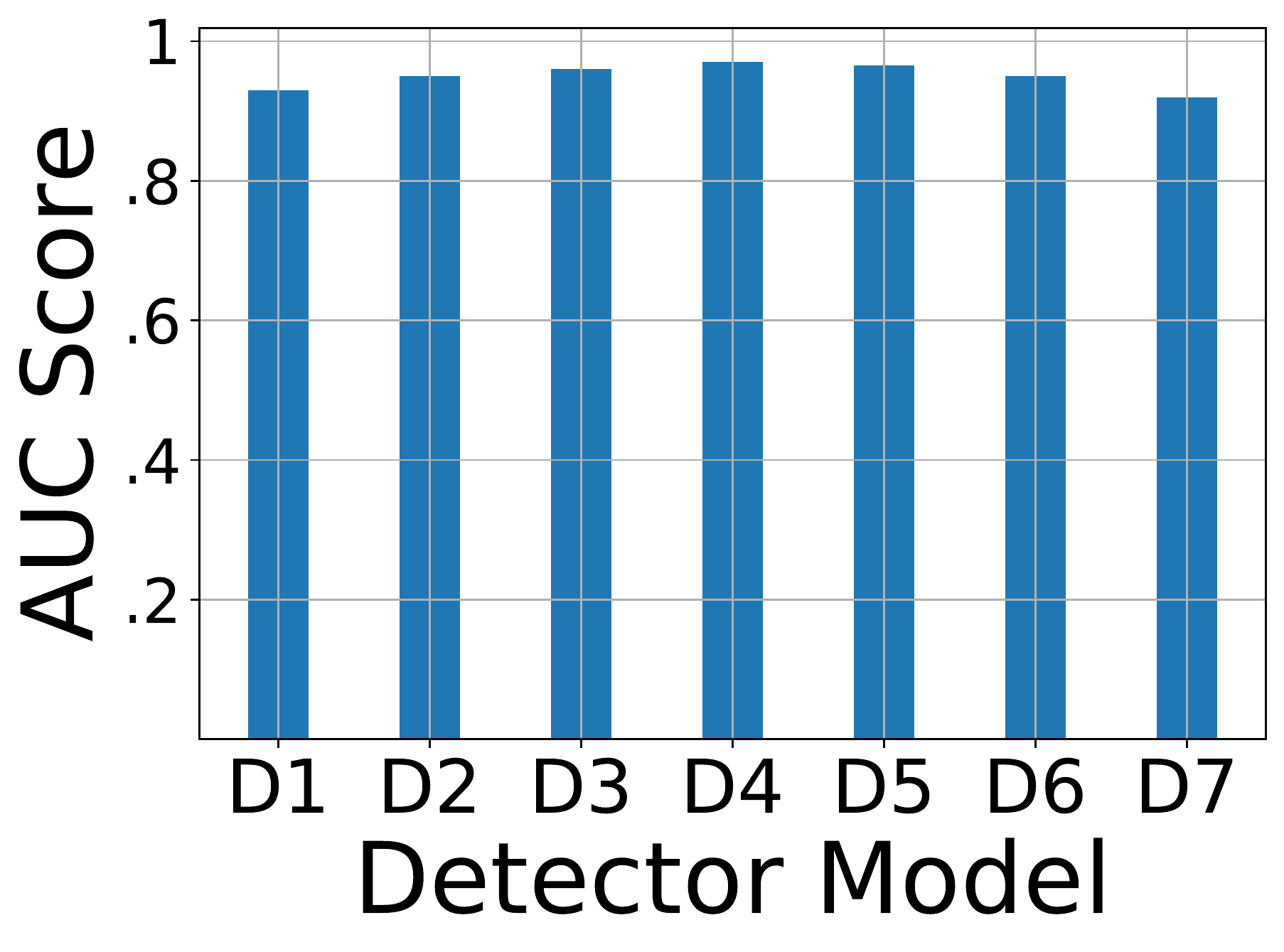}
         \caption{}
         \label{lambda_select}
     \end{subfigure}
    \caption{a) AUC scores corresponding to different strengths of PGD attacks- PGD2: PGD(2,1,10), PGD4: PGD(4,2,10), PGD8: PGD(8,4,10), and PGD16: PGD(16,4,10) for five different $K$ values. b) Accuracy values corresponding to the natural data for five different $K$ values. c) AUC scores for PGD(4,2,10) attack across 7 different models trained using different $\lambda_a$ values. }
    \label{lwise_training}
    \vspace{-4mm}
\end{figure}

\subsection{Effect of Choosing the Threshold Percentile}
\label{threshold choice}
In this section, we show how adversarial detection is affected by the choice of $\mathcal{E}_{Th}$. The value $\mathcal{E}_{Th}$ is chosen as the $K^{th}$ percentile of the $\mathcal{E}_{s_{nat}}$ distribution. Fig. \ref{K_select} shows the AUC scores for four different strengths of PGD attacks across five different $K$ values corresponding to CIFAR100 dataset. It is observed that lower values of $K$ yield higher AUC scores for weak attacks (such as PGD2). However, the AUC scores for stronger attacks are lower. Additionally, at lower $K$ values, the accuracy (See Section \ref{auc_error_accuracy}) is lower as seen in Fig. \ref{K_select_error}. This is because at low $K$ values, higher number of clean data samples are mis-classified as adversarial. At extremely high values of $K$ ($K$= 98), the AUC scores for weak attacks significantly fall. However, a higher accuracy is achieved. In both Fig. \ref{K_select} and Fig. \ref{K_select_error}, $S_T$=ResNet18 and $S_V$=Mobilenet-v2 and $M_C$=VGG19. Evidently, choosing the $K$ value is a trade-off between the detection of weaker attacks and accuracy. Therefore, $K$ is a design choice that can be set according to the designer's priority. In this work, we choose $K$=95 for all our experiments.   


\subsection{Selecting $\lambda_n$ and $\lambda_a$}
$\lambda_n$ and $\lambda_a$ are hyperparameters that denote the desired natural and adversarial energies for a particular layer $l$ in the detector. We show how the choice of $\lambda_n$ and $\lambda_a$ affects the detector performance. For this, we train 7 different detectors with different $\lambda_a$ values. D1(0.3,0.7,1.7) (with $\lambda_{a}$=0.3, 0.7, 1.7 for layer 1, 2, 3, respectively), D2(0.5,0.9,1.9), D3(0.7,1.1,2.1), D4(0.9,1.3,2.3) and D5(1.1,1.5,2.5), D6(1.3,1.7,2.7), D7(1.5,1.9,2.9). For all the models, $\lambda_n$= 0.1. As seen in Fig. \ref{lambda_select} detection capability increases from D1 to D4 as the $\lambda_{a}$ increases and decreases beyond M4. However, the changes in AUC scores for different sets of $\lambda_a$ values is marginal. We find that all detector models (D1-D7) are good detectors for CIFAR10, CIFAR100 and TinyImagenet datasets (see Supplementary Material).


\section{Results}
\label{results_section}
\subsection{Experimental Setup}
We use three different image datasets for our experiments: CIFAR10 \cite{krizhevsky2009learning}, CIFAR100 \cite{krizhevsky2009learning}, and TinyImagenet \cite{deng2009imagenet}. Both the training and validation data are scaled between (0,1) in all our experiments. For generating adversarial attacks, we use the torchattacks library \cite{kim2020torchattacks}. 

For all our experiments, we use a 3-layered detector with the following architecture: Conv(3,8)- ReLU- Conv(8,16)- ReLU- Conv(16,32). For training the 3-layered detector, the LES training contains 3 stages. The LES training is performed with adversarial data created by PGD(8,4,10) attack and model $S_T$. For the CIFAR100 and CIFAR10 dataset, LES training is carried out with learning rates 0.005, 0.005 and 0.001, respectively in each stage. While for the TinyImagenet dataset, the learning rates correspond to 0.003, 0.001 and 0.001). During LES training, we use $\lambda_n$= 0.1 while $\lambda_a$= (0.9, 1.3, 2.3). The LES training is carried out for 500 epochs in each stage. Due to the small detector network size, the training time is significantly small (1 to 2 GPU hours for the 3-layered detector). Additionally, models $S_T$, $S_V$ and $M_C$ are trained on the respective datasets using the SGD algorithm. All experiments are conducted using Pytorch 1.5 platform with Nvidia rtx2080ti GPU backend.

\subsection{Performance on Different Adversarial Attacks}
\begin{table*}[t]

\resizebox{1\textwidth}{!}{
    \begin{tabular}{l|c|r|ccccccccc}\hline
         Detector Model&\multicolumn{2}{c}{Attacks $\rightarrow$}  &\: FGSM &\: PGD4 &\: PGDL2 &\: C$\&$W &\: SQR &\: APGD &\: AUTO &\: GN &\: FAB \\
         \hline
         \multirow{5}{*}{\begin{tabular}{l} \textbf{Black-Box}\\ \\$S_T$: ResNet18 \\ \textbf{$S_V$: MobileNet-v2} \\ $M_C$: VGG19 \\ \:\end{tabular}} & \multirow{2}{*}{\begin{tabular}{c}No\\Detector\end{tabular}} & Error &\: 66 &\: 54 &\: 91 &\: 63 &\:88 &\: 53 &\: 69 &\: 77 &\: 78\\ \cline{4-12}
         & & Accuracy &\cellcolor{black!8} &\cellcolor{black!8} \cellcolor{black!8}&\cellcolor{black!8} &\cellcolor{black!8} & 59\cellcolor{black!8} &\cellcolor{black!8} &\cellcolor{black!8} & \cellcolor{black!8}&\cellcolor{black!8} \\ \cline{2-12}
         &\multirow{2}{*}{\begin{tabular}{c}With\\Detector\end{tabular}} & Error & 0 & 0 & 0 & 0 & 23 & 28 & 28 & 0 & 72 \\ \cline{4-12}
         & & Accuracy &\cellcolor{black!8} &\cellcolor{black!8} \cellcolor{black!8}&\cellcolor{black!8} &\cellcolor{black!8} & 56\cellcolor{black!8} &\cellcolor{black!8} &\cellcolor{black!8} & \cellcolor{black!8}&\cellcolor{black!8}  \\ \cline{2-12}
         & \multicolumn{2}{c|}{AUC Score} & 0.98 & 0.97 & 0.98 & 0.79 & 0.79 & 0.78 & 0.79 & 0.98 & 0.5\\ \hline
         \multirow{5}{*}{\begin{tabular}{l} \textbf{White-Box}\\ \\$S_T$: ResNet18\\ \textbf{$S_V$: VGG19} \\ $M_C$: VGG19 \\ \:\end{tabular}} & \multirow{2}{*}{\begin{tabular}{c}No\\Detector\end{tabular}} & Error & 82 & 95 & 95 & 100 & 99 & 92 & 100 & 77 & 99 \\ \cline{4-12}
         & & Accuracy &\cellcolor{black!8} &\cellcolor{black!8} \cellcolor{black!8}&\cellcolor{black!8} &\cellcolor{black!8} & 59\cellcolor{black!8} &\cellcolor{black!8} &\cellcolor{black!8} & \cellcolor{black!8}&\cellcolor{black!8} \\ \cline{2-12}
         &\multirow{2}{*}{\begin{tabular}{c}With\\Detector\end{tabular}} & Error & 0 & 0 & 19 & 39 & 39 & 39 & 39 & 0 & 92 \\ \cline{4-12}
         & & Accuracy &\cellcolor{black!8} &\cellcolor{black!8} \cellcolor{black!8}&\cellcolor{black!8} &\cellcolor{black!8} & 56\cellcolor{black!8} &\cellcolor{black!8} &\cellcolor{black!8} & \cellcolor{black!8}&\cellcolor{black!8} \\ \cline{2-12}
         & \multicolumn{2}{c|}{AUC Score} & 0.98 & 0.97 & 0.98 & 0.79 & 0.8 & 0.78 & 0.8 & 0.98 & 0.5 \\ \hline
    \end{tabular}}
    
    \caption{Table showing the AUC score, error and accuracy value for a single detector model subject to white-box and black-box attacks. Note, accuracy value is detector specific and does not depend on the type of attack.}
    \label{diff_attacks_err}
\end{table*}

\begin{wraptable}{r}{0.7\textwidth}
\centering
\caption{Table showing the AUC scores for different detectors trained on CIFAR100. The detectors have high detection capability for different adversarial attacks irrespective of the architecture of the $S_V$ and $S_T$ models.}
\label{tab:model_agnostic}
\resizebox{\textwidth}{!}{
\begin{tabular}{lcc|cc|cc}
& \multicolumn{2}{c}{Detector 1} & \multicolumn{2}{c}{Detector 2} & \multicolumn{2}{c}{Detector 3}\\ \hline
\multirow{2}{*}{Attacks $\downarrow$} & \multicolumn{2}{c|}{$S_T$ MobileNet-V2}& \multicolumn{2}{c|}{$S_T$ ResNet18} & \multicolumn{2}{c}{$S_T$ VGG16} \\
 & \:$S_V$ ResNet18 &\: VGG16 \:& \: $S_V$ VGG16 & \:MobileNet-V2 \:&\: $S_V$ ResNet18 &\: MobileNet-V2  \\
\hline
\multirow{14}{*}{\begin{tabular}{l|}
     FGSM \cite{kurakin2016adversarial} \:\\ BIM \cite{kurakin2016adversarial} \:\\ PGD8 \cite{madry2017towards} \: \\ PGD4 \cite{madry2017towards} \: \\PGDL2 \cite{madry2017towards} \: \\ FFGSM \cite{wong2020fast}  \:\\ TPGD \cite{zhang2019theoretically} \:\\ MIFGSM \cite{dong2018boosting} \:\\ DIFGSM \cite{xie2019improving}\:\\ C\&W  \cite{carlini2017towards} \: \\ SQR \cite{andriushchenko2020square} \\ APGD \cite{croce2020reliable} \:\\ AUTO \cite{croce2020reliable} \:\\ GN \:\end{tabular}} & \multirow{14}{*}{\begin{tabular}{c}0.98\\ 0.98\\ 0.98\\ 0.97\\ 0.87\\ 0.98\\ 0.98\\ 0.98\\ 0.97\\ 0.75\\ 0.75\\ 0.74\\ 0.75\\ 0.98 \end{tabular}} & \multirow{14}{*}{\begin{tabular}{c}0.98\\ 0.97\\ 0.98\\ 0.96\\ 0.83\\ 0.98\\ 0.97\\ 0.97\\ 0.97\\ 0.78\\ 0.77\\ 0.74\\ 0.77\\ 0.98 \end{tabular}} & \:\: \multirow{14}{*}{\begin{tabular}{c}0.98\\ 0.98\\ 0.98\\ 0.97\\ 0.85\\ 0.98\\ 0.98\\ 0.98\\ 0.98\\ 0.77\\ 0.77\\ 0.75\\ 0.77\\ 0.98 \end{tabular}} & \multirow{14}{*}{\begin{tabular}{c}0.98\\ 0.98\\ 0.98\\ 0.97\\ 0.98\\ 0.98\\ 0.98\\ 0.98\\ 0.98\\ 0.79\\ 0.79\\ 0.78\\ 0.79\\ 0.98 \end{tabular}} & \multirow{14}{*}{\begin{tabular}{c}0.98 \\ 0.98 \\ 0.98 \\ 0.97 \\ 0.92 \\ 0.98 \\ 0.98 \\ 0.98 \\ 0.98 \\ 0.75 \\ 0.75 \\ 0.74 \\ 0.75 \\ 0.98 \end{tabular}} & \multirow{14}{*}{\begin{tabular}{c}0.98\\ 0.98\\ 0.98\\ 0.97\\ 0.98\\ 0.98\\ 0.98\\ 0.98\\ 0.98\\ 0.79\\ 0.79\\ 0.78\\ 0.79\\ 0.98 \end{tabular}} \\
 &   &  & \:\:  &  \\
 &   &  & \:\:  & \\
 &   &  & \:\:  & \\
 &   & & \:\:  & \\
 &   &  & \:\: & \\
 &   &  & \:\:  & \\
 &   &  & \:\:  &\\
 &   &  & \:\:  & \\
 &   &  & \:\:  & \\
 &   &  & \:\:  & \\
 &  &  & \:\:&  \\
 &   &  & \:\:  &\\
 &   &  & \:\: & \\
  \hline 
\end{tabular}}
\end{wraptable}\vspace{-3mm}

In Table \ref{diff_attacks_err}, we consider an LES-trained detector on the CIFAR100 dataset with $S_T$=ResNet18. The \\classifier model $M_C$= VGG19. We consider white-box and black-box scenarios for validating the performance of the given detector. In white-box attack, the attacker has complete knowledge of the classifier model. While in black-box attack, the attacker does not have any information about the classifier model. To this effect, we use $S_V$: MobileNet-v2 and VGG19, for black-box and white-box attacks, respectively. Note, the premise of this paper is to train an adversarial detector and perform adversarial detection without any classifier model information. The white-box attack scenario considered here pertains only to the generation of attacks (\ie $S_V$: VGG19) and not during detection or LES training.  

Evidently, the detector has a high AUC score across a wide range of gradient, score and gaussian noise attacks (attack parameters shown in Supplementary Material). Consequently, adding the detector significantly lowers the the error (see Section \ref{auc_error_accuracy}) value of the classifier model compared to the ``No Detector" case. Further, the reduction is substantially higher in case of white-box attacks compared to black-box attacks. However, we find that adding the detector leads to a 3\% drop in accuracy (see Section \ref{auc_error_accuracy} and Section \ref{threshold choice}) compared to the ``No Detector" case. 

Note, that the detector fails to detect decision-based attacks like the FAB attack. This is because FAB attacks add minimal perturbations to the natural input which does not change the $\mathcal{E}_{adv}$ value significantly. Consequently, a high error value is observed.

Interestingly, the detector shown in Table \ref{diff_attacks_err} demonstrates model agnostic behavior. For example, the detector can identify adversarial inputs generated using models $S_V$= VGG19 and $S_V$= MobileNet-v2. To further illustrate this model agnostic property, we create three detectors- Detector 1: $S_T$= \\MobileNet-v2, Detector 2: $S_T$= ResNet18 and Detector 3: $S_T$= VGG16. All the detectors are trained using LES training. Table \ref{tab:model_agnostic} shows the AUC scores of the detectors under adversarial attacks generated using $S_V$ models that have entirely different network architectures from the model $S_T$ used in the training. Evidently, the detectors have significantly high performance over different adversarial attacks and are agnostic to the $S_T$ and $S_V$ models used during training and attack generation, respectively. Note, AUC scores for FAB attack have not been shown as it is known that our detection approach cannot identify FAB attacks. Similar observations can be made for the CIFAR10 and TinyImagenet datasets and have been shown in  Supplementary Material.
\subsection{Comparison with Prior Works}
\begin{table*}[t]
\centering
\caption{{Table comparing the AUC scores, memory and computation overhead of our method with prior state-of-the-art adversarial detection works. AUC scores that are not reported by prior works have not been shown.}}
\label{tab:comparison}
\resizebox{\textwidth}{!}{%
\begin{tabular}{lccccccc}
\toprule
\multirow{2}{*}{Work} & \multirow{2}{*}{Dataset} & \multicolumn{2}{c}{Weak Attacks} & \multicolumn{2}{c}{Strong Attacks} & \multicolumn{1}{c}{\multirow{2}{*}{\begin{tabular}[c]{@{}c@{}}Number of\\ Parameters\end{tabular}}} & \multicolumn{1}{c}{\multirow{2}{*}{\begin{tabular}[c]{@{}c@{}}Number of\\ Operations\end{tabular}}} \\ \cline{3-6}
 &  & FGSM $\epsilon$: 4/255 & PGD $\epsilon$: 4/255 & FGSM $\epsilon$: 16/255 & PGD $\epsilon$: 16/255 & \multicolumn{1}{c}{} & \multicolumn{1}{c}{} \\ \hline
Metzen et al. \cite{metzen2017detecting} & CIFAR10 & 1 & 0.96 & Not Reported & Not Reported & $500k$ & $1.4M$ \\
Yin et al. \cite{yin2019gat} & CIFAR10 & Not Reported & Not Reported & Not Reported & 0.953 & $ 213k$ & $ 6.04M$ \\
Moitra et al. \cite{moitra2021detectx} & CIFAR10 & 0.85 & 0.88 & 0.98 & 0.895  & $500k$ & $1.7M$ \\
Sterneck et al. \cite{sterneck2021noise} & CIFAR10 & 0.99 & 0.998 & 1 & 1 & $525k$ & $4.7M$ \\
Xu et al. \cite{xu2017feature} & CIFAR10 & 0.208 & 0.505 & Not Reported & Not Reported & $ 10G$ & $ 10G$ \\
Gong et al. \cite{gong2017adversarial} & CIFAR10 & 0.003 & Not Reported & 1 & Not Reported &$64k$ & $477k$ \\ 
\textbf{Ours} & \textbf{CIFAR10} & \textbf{0.97} & \textbf{0.98}  & \textbf{0.98} & \textbf{0.98} & $ \textbf{5.9k}$ & $ \textbf{500k}$ \\
\hline
Sterneck et al. \cite{sterneck2021noise} & CIFAR100 & 0.95 & 0.99 & 0.99 & 1  & \multicolumn{1}{c}{$525k$} & \multicolumn{1}{c}{$4.7M$} \\
Moitra et al. \cite{moitra2021detectx} & CIFAR100 & 0.6 & 0.64 & 0.98 & 0.99 & \multicolumn{1}{c}{$500k$} & \multicolumn{1}{c}{$1.7M$} \\ 
\textbf{Ours} & \textbf{CIFAR100} & \textbf{0.96} & \textbf{0.97} & \textbf{0.98} & \textbf{0.98}  & \multicolumn{1}{c}{$\textbf{5.9k}$} & \multicolumn{1}{c}{$\textbf{500k}$} \\
\hline
Moitra et al. \cite{moitra2021detectx} & TinyImagenet & 0.52 & 0.56 & 0.84 & 0.65 & \multicolumn{1}{c}{$500k$} & \multicolumn{1}{c}{$1.7M$} \\ 
\textbf{Ours} & \textbf{TinyImagenet} & \textbf{0.97} & \textbf{0.98} & \textbf{0.98} & \textbf{0.98}  & \multicolumn{1}{c}{$\textbf{5.9k}$} & \multicolumn{1}{c}{$\textbf{147k}$} \\
\hline
\end{tabular}%
}
\end{table*}\vspace{-3mm}
Table \ref{tab:comparison} compares the performance and cost-effectiveness of our proposed method with prior state-of-the-art adversarial detection works. For comparison, detectors for different datasets are created using LES training with $S_T$=ResNet18. $S_V$= MobileNet-v2. Clearly, we achieve comparable detection performance compared to prior adversarial detection approaches. It must be noted that our work does not aim to outperform prior adversarial detection works. Instead, the striking feature of our approach is to overcome the dependence of prior adversarial detection works on the classifier model for training and adversarial detection. 

\noindent\textbf{Computational overhead:} Besides high adversarial detection, our method requires  10-100x less number of operations and parameters for adversarial detection compared to all the prior works. Here, the number of operations is estimated as total number of dot-product computations across all layers of the detector. Note, prior works use large number and size of detectors. Further, as their approaches are classifier model dependent, their detectors are always implemented with the classifier model leading to a high memory and computation overhead. The low memory and computation overhead makes our approach suitable for deployment in resource constrained systems.

\subsection{LES-Training with Limited Training Data}
\begin{wrapfigure}{h}{0.38\textwidth}
    
    \vspace{-9mm}
    \centering
    \begin{tabular}{c}
    \includegraphics[width=1\linewidth]{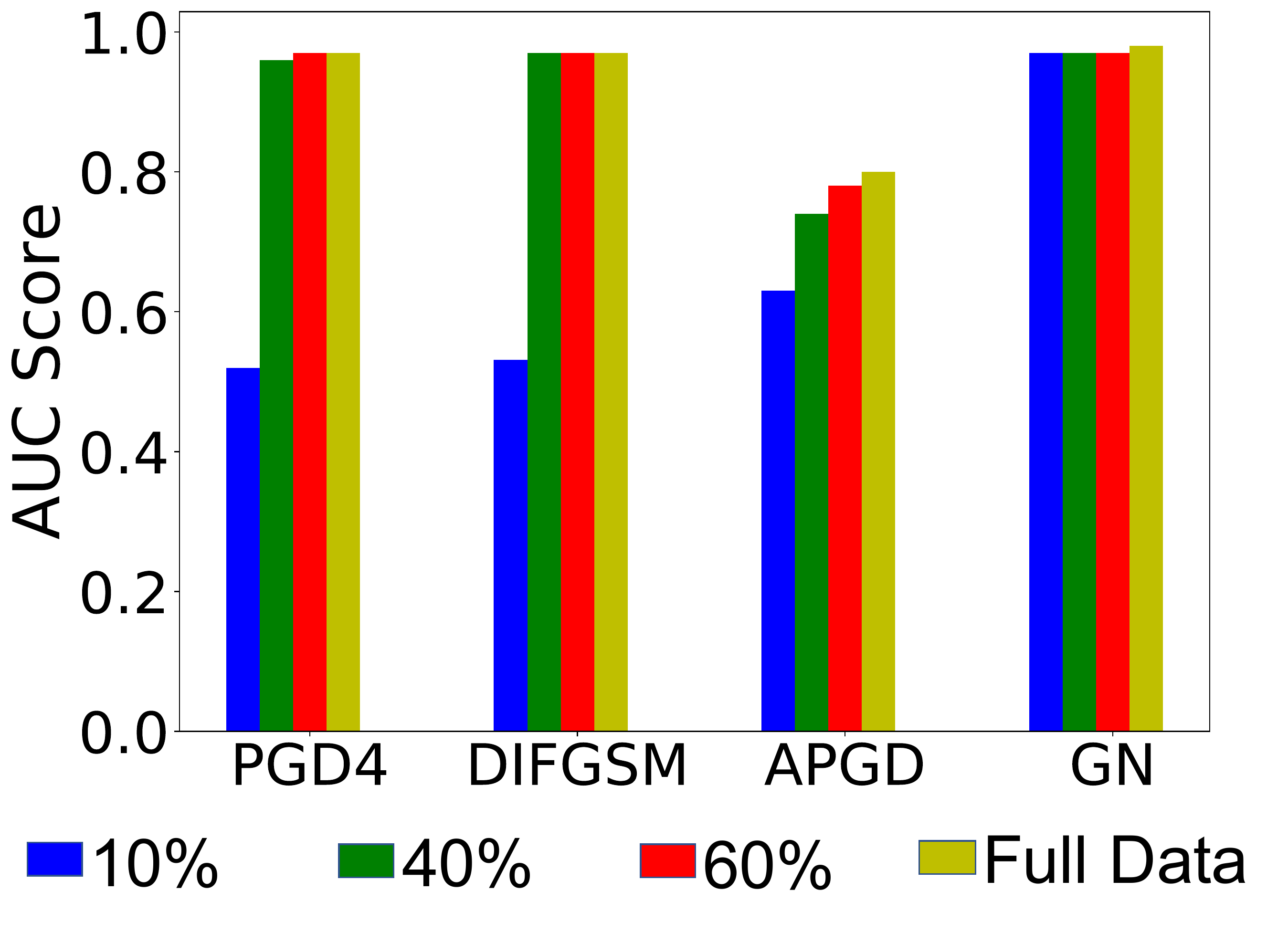}  \\
    \end{tabular}
    \vspace{10mm}
    \caption{AUC scores corresponding to different adversarial attacks for detectors created using 10\% , 40\% and 60\% of the CIFAR100 dataset.}\vspace{-15mm}
    \label{lim_data}
    
\end{wrapfigure}
In this section we evaluate the performance of different detectors created with limited access to the actual dataset. For this, we create 3 detectors trained on subsets of size 1)10\% 2)40\% and 3)60\% of the CIFAR100 dataset. The subsets are created by random sampling from the actual dataset. For reference, we also show the performance of a detector trained on the full dataset. In all cases, the detectors are created with $S_T$= ResNet18 and tested on attacks generated using $S_V$=MobileNet-v2 networks. It can be observed that detectors trained with just 40\% of the training data can achieve AUC scores comparable with the detector trained on the full data. Interestingly, for some attacks (such as GN) merely 10\% of the training data is sufficient for achieving a high performance. Please refer to Supplementary Material for similar results on the TinyImagenet and CIFAR10 datasets.

\subsection{Transferability Across Datasets}

In this section, we explore the following question: \textit{Can a detector trained on dataset A (source dataset) be used to detect adversaries from another dataset B (target dataset)?}.
\noindent\textbf{Methodology: } LES training is performed to create a detector network on the source dataset. Following this, a sample dataset $s_{nat, target}$ is sampled from the target dataset. Through our experiments, we find that a size of 200 samples for $s_{nat, target}$ data is sufficient for transferability to the target dataset. The $s_{nat, target}$ and LES-trained detector is used to generate the $\mathcal{E}_{s_{nat,target}}$ distribution. $\mathcal{E}_{Th}$ is chosen as the 95th percentile of the $\mathcal{E}_{s_{nat,target}}$.
\begin{figure*}[t] 
\centering
     \includegraphics[width=0.8\textwidth]{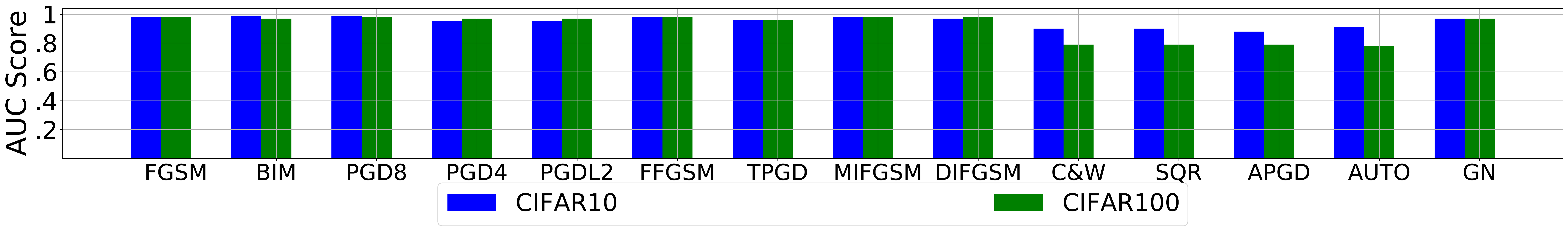} 
     \caption{AUC scores for a detector trained with source dataset TinyImagenet and transferred to target datasets CIFAR10 and CIFAR100.}\vspace{-4mm}
     \label{transfer_tiny}
\end{figure*}
\noindent\textbf{Transferability of detectors trained on TinyImagenet.} We train a detector D-Tiny with source dataset= TinyImagenet (shown in Fig. \ref{transfer_tiny}). The detector is trained using LES training with $S_T$= ResNet18 network. D-Tiny is transferred to target datasets-  CIFAR100 and CIFAR10 using the methodology explained above. Here, $S_V$= MobileNet-v2. Evidently, D-Tiny being trained on the TinyImagenet dataset transfers well across both CIFAR10 and CIFAR100 datasets. For example, D-Tiny achieves similar performance against attacks on the CIFAR100 dataset as Detector1, Detector 2 and Detector 3 from Table \ref{tab:model_agnostic}.

\begin{wrapfigure}{h}{0.32\textwidth}
    \vspace{-5mm}
    \centering
    \includegraphics[width=1.05\linewidth]{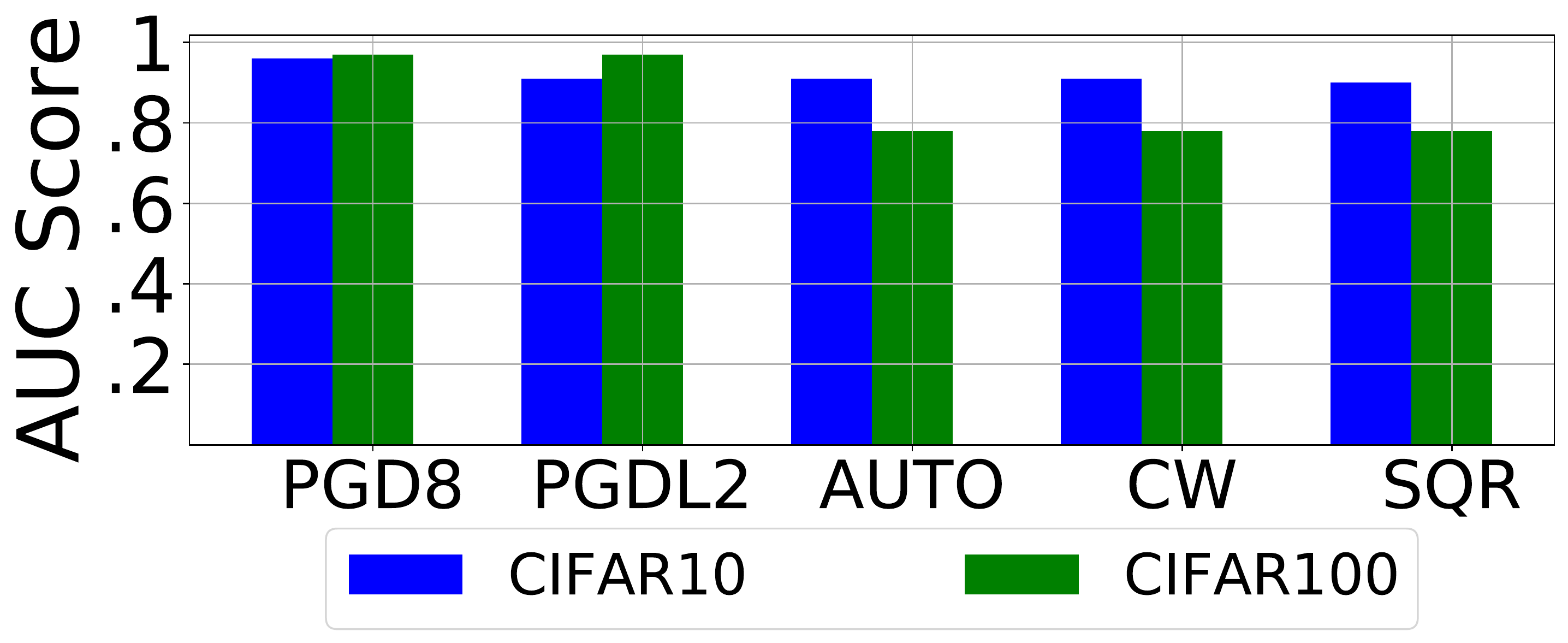}     \vspace{1mm}
    \caption{A detector trained on 40\% of TinyImagenet successfully transfers to CIFAR100 and CIFAR10. }
    \label{lim_data_transfer}
    \vspace{-6mm}
\end{wrapfigure}
Interestingly, even detectors trained on smaller dataset like CIFAR100 can transfer reasonably to a larger dataset such as TinyImagenet (see Supplementary Material).

    

In Fig. \ref{lim_data_transfer}, we evaluate the transferability of a detector D-Tiny40 trained on 40\% of source dataset TinyImagenet and transferred to target datasets CIFAR100 and CIFAR10.  We find that even detectors trained with limited access to the TinyImagenet dataset can transfer successfully. Similar experiments with other datasets are shown in the Supplementary Material.
\section{Conclusion}
In this work, we propose a classifier model independent energy separation-based adversarial detection method that does not require access to the classifier model. Further, we achieve comparable detection performance on a wide range of adversarial attacks at an extremely small memory and computational overhead compared to prior works. Moreover, we show that our method achieves requires lesser data compared to prior works while achieving significant performance. Further the detector is also transferable across different datasets. However, our method entails few limitations. Although our detection is classifier model agnostic, it still requires access to the data. Although the data dependency of our approach is lower compared to prior works, an interesting future direction can be using synthetic data during LES training and transferring to real-world data without accessing any partial data. This will make our detection approach more robust. Further, our approach, while successfully detecting a suite of attacks, fails to detect decision-based adversarial attacks. Future works can focus on finding more sophisticated energy functions to detect such attacks.

\clearpage
%
%
\bibliographystyle{splncs04}
\bibliography{eccv_paper.bib}
\end{document}